# Text Detection on Technical Drawings for the Digitization of Brown-field Processes

Tobias Schlagenhauf[a,*], Markus Netzer[a], Jan Hillinger[a]

[a]*Karlsruhe Institute of Technology, wbk Institute of Production Science, Kaiserstraße 12, 76131 Karlsruhe, Germany*

**Abstract**

This paper addresses the issue of autonomously detecting text on technical drawings. The detection of text on technical drawings is a critical step towards autonomous production machines, especially for brown-field processes, where no closed CAD-CAM solutions are available yet. Automating the process of reading and detecting text on technical drawings reduces the effort for handling inefficient media interruptions due to paper-based processes, which are often today's quasi-standard in brown-field processes. However, there are no reliable methods available yet to solve the issue of automatically detecting text on technical drawings. The unreliable detection of the contents on technical drawings using classical detection and object character recognition (OCR) tools is mainly due to the limited number of technical drawings and the captcha-like structure of the contents. Text is often combined with unknown symbols and interruptions by lines. Additionally, due to intellectual property rights and technical know-how issues, there are no out-of-the box training datasets available in the literature to train such models. This paper combines a domain knowledge-based generator to generate realistic technical drawings with a state-of-the-art object detection model to solve the issue of detecting text on technical drawings. The generator yields artificial technical drawings in a large variety and can be considered as a data augmentation generator. These artificial drawings are used for training, while the model is tested on real data. The authors show that artificially generated data of technical drawings improve the detection quality with an increasing number of drawings. Moreover, the trained model can be used to reliably detect text on real technical drawings. The authors were able to show that the model performs better than the current state-of-the-art models. The code is available at: https://github.com/2Obe/Text-Detection-on-Technical-Drawings

## 1. Introduction

In industrial brown-field environments, there are often no closed-loop CAD-CAM solutions. This leads to unintentional media disruptions, as technical drawings are often only available in paper form. To enable the use of autonomous production machines, information on technical drawings in brown-field environments often has to be extracted manually by the employees to be then interpreted and fed back into a digital processing device. These media discontinuities are inefficient but cannot be directly prevented due to the still prevalent brown-field conditions. All the more, procedures are needed that can fix these media discontinuities. To implement this in the context of technical drawings, an Optical Character Recognition (OCR) model is needed that is able to analyze the drawings and return the content in digital form. However, so far, no OCR model is known to have been designed for this task and enable recognition of not only text but also symbols on technical drawings. Existing OCR models deal with use cases such as captcha code and license plates. However, these existing models are not suitable for technical drawings. The main reason is that they do not recognize special characters, texts with specific properties, or occurring overlays of certain elements, as these are specific to technical drawings. The task of recognizing text in technical drawings is compounded by another challenge. Since technical drawings contain critical company information, they are usually not freely accessible, which is why there are no freely accessible data sets that can be used to train models. However, the structure of technical drawings is the same across companies. In this paper, the authors present a new approach to detecting dimensions, positional tolerances, and shape tolerances under the challenge of the severely limited availability of real engineering drawings. Our contribution is to

- develop a generator to create an artificial dataset for training a text recognition approach on technical drawings,
- investigate an approach to the reliable detection of dimensions, positions, and shape tolerances on technical drawings.

## 2. Related Work

The commercial use of OCR applications began in the early 1950s [1]. Early works by [2] [3] rely mainly on the application of algorithms such as the Stroke Width Transform and Maximally Stable Extremal Regions. In the mid-2010s, new approaches were added due to Deep Learning. One of the most significant approaches is the Fast Region Based Convolutional Neural Network (Fast-RCNN). This uses a Convolutional Neural Network for both the proposals and the detection itself. Compared to existing approaches, this has resulted in better accuracy and increased speed [4]. Recently, the use of OCR employing deep learning techniques to specific tasks has been investigated. [5] [6] present general deep learning-based methods dedicated to the detection and recognition of text in different real-world contexts. [7] [8] also deal with the detection of characters on Piping & Instrumentation Diagrams (P&ID), whereby the approaches partly show weaknesses in the detection at certain line elements. A similar approach can be found for construction drawings [9]. In summary, there are no approaches at the current state of the art that allow a reliable detection of characters on industrial technical drawings. Also, no public dataset for specific training of the model for technical drawings is known so far. Based on the gaps in the state of the art, we develop an approach that is able to reliably detect text on technical drawings and requires only a small amount of real-world training data with help through a generator.

## 3. Method

Text recognition requires an OCR model. Classically, this consists of a detector, which localizes text on an image, and a recognizer, which deciphers the localized text. In addition, data is required to train both parts. To meet the challenge of detecting text in technical drawings in the context of limited data sets, we proceed in three steps. First, a so-called generator is designed, which is able to generate further artificial technical drawings on the basis of a small number of real technical drawings. This generator is used to generate training data for both the detector and the recognizer and to train the two submodels. Once the training is complete, the actual text recognition on technical drawings follows. The technical drawings are first given to the model for text detection. If places with text on the technical drawing are localized, these are isolated from the remaining drawing. The locations with text are then fed to a text recognition model to decipher the respective text. To demonstrate the effectiveness of the approach, experiments are conducted with generated and real technical drawings in the form of images. Figure 1 illustrates the interaction of the individual components, which will be discussed in the following.

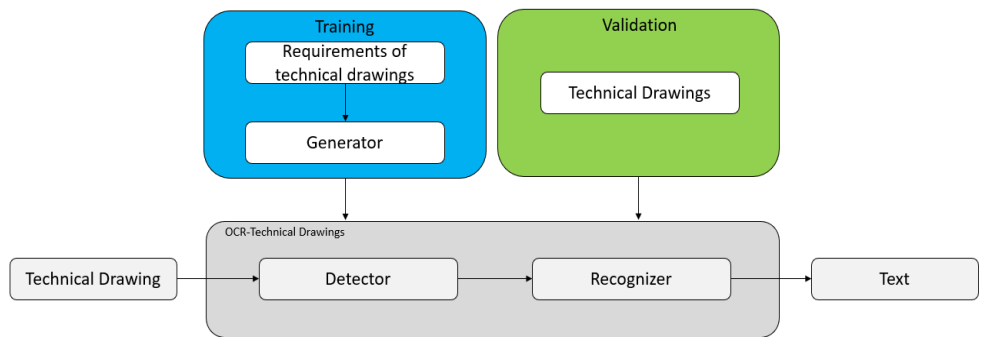

Figure 1. Visualization of the presented approach.

*3.1 Generation of Technical Drawings by Use of a Generator*

Since annotated images of technical drawings are necessary for the training of the detection and recognition models, a field of tension exists. Although models that are potentially able to process technical drawings are available in the state of the art, the necessary amount of data is lacking. In contrast to real-world objects (e.g. dogs, cats, humans), whose characteristics do not follow any fixed rules, the structure of technical drawings is standardized. Due to the structure of technical drawings, this domain knowledge can be fed into the generator for data generation. The generator is thus able to generate images that represent artificial technical drawings in order to train the approach for text recognition on technical drawings. For the detector, the generator generates images that follow fixed frameworks but have different characteristics. To ensure the greatest possible variety of images, two different image types are generated. Image type 1 is used for the realism of real technical drawings. This means that labeled geometry is generated on these images, just like on real technical drawings. To further enable the model to distinguish objects (e.g. dimensions or tolerances) from background, an image type 2 is generated. The two image types are illustrated in Figure 2. The background on technical drawings is everything contained in the image that does not represent the dimensions or tolerances being searched for. Dimensions have a high density of black pixels compared to other regions of engineering drawings. However, dimensions are not the only objects of technical drawings with such pixel densities. Thus, image type 2 is used to distinguish dimensions from other drawing elements such as hatches, threads, bearings, and text boxes to reduce misinterpretation of these elements. Image type 1 contains the information that should be learned, image type 2 those that should not be learned erroneously as objects of interest. Besides the generation of the images, the augmentation and annotation of the images is performed simultaneously. Thus, the generated images of both image types are immediately available for training. The two described image types have varying dimensions, so that the approach can handle drawings of different sizes. The image width is randomly chosen between 1200 and 1500 pixels. The image height depends on the image width and follows the standard format of 1:√2. On the generated images, the different objects have different sizes. The average size of the dimensions is 45x25 pixels, the size of the shape and position tolerances is 100x25 pixels. In order to guarantee the variety of the generated images, there are different characteristics for the two image types. Image type 1 has varying, randomly generated part geometries. This results at the same time in a varying labeling. To obtain the greatest variety of dimensions, the lengths of the dimensions are independent of the actual size of the geometry and are randomly generated with a number from 0 to 300 pixels. In addition, the labeled part is randomly arranged on the image to cover any region in which a part can occur. In order to consider different image qualities of real drawings, random augmentations of sharpness, contrast and brightness of the generated images of image type 1 are performed. For image type 2, 150 different elements (e.g. bearings, drillings, hatchings) cut out from 9 real drawings are used. Real elements from drawings are used, because with the help of these elements, images can be generated easily. Another advantage is that these elements are at the same time the connection to real drawings and thus realistic expressions can be covered. These elements are randomly placed in random number, random selection, on the generated image of image type 2. The dimensions and tolerances to be trained are subsequently placed randomly on the image in horizontal or vertical orientation in random expression. This achieves highly individual arrangements of object and background, resulting in a high diversity of images. Image type 1 has between 10 and 16 dimensions and tolerances per image. Image type 2 contains 40 dimensions and tolerances per image. Image type 3 was created for the recognizer of the approach. The size of the image is 50x35 pixels. These are images with simple character sequences, which consist of a certain selection of characters found on technical drawings. This selection consists of the numbers 0-9, period and comma, the letters F, G, H, K, M frequently occurring on technical drawings and the symbols +, -, ∓ and ∅. On all generated images of image type 3, the sequence of digits consists of 2 to 4 random digits of the described selection. Image types 1 and 2 thus ensure that relevant text passages are reliably detected. Image type 3 is used to reliably recognize the text within relevant areas.

### 3.2 Object Detection

After training images have been generated, the detection model is selected and parameterized. As detection model, a Faster-RCNN model was selected for the object detection part, because it has good performance on real images [10]. For this purpose, the freely available Faster-RCNN ResNet50 model from Tensorflow was used [11]. This is a region proposal network that generates proposals for the presence of certain objects. These proposals are then passed through a classifier and regressor to ultimately decide if it really is a searched object. More information can be found at [10]. By means of this model, a so-called transfer learning is applied. In this process, weights that have already been trained using the Common Objects in Context (COCO) data set are used and only the last layers of the neural network are trained again. In preliminary experiments, transfer learning was identified as an effective method. This is also confirmed in the following results. The selected model transforms all training images to the size of 640x640px. To recognize the objects searched in this paper, the model was trained on 3 classes. "dimensions horizontal", "dimensions vertical", and "shape and position tolerances". The following values were identified as effective parameters: Number of Anchors: 24, Number of Iterations: 20,000, Learning Rate: 0.001. The detector is trained with different data sets depending on the different experiments presented.

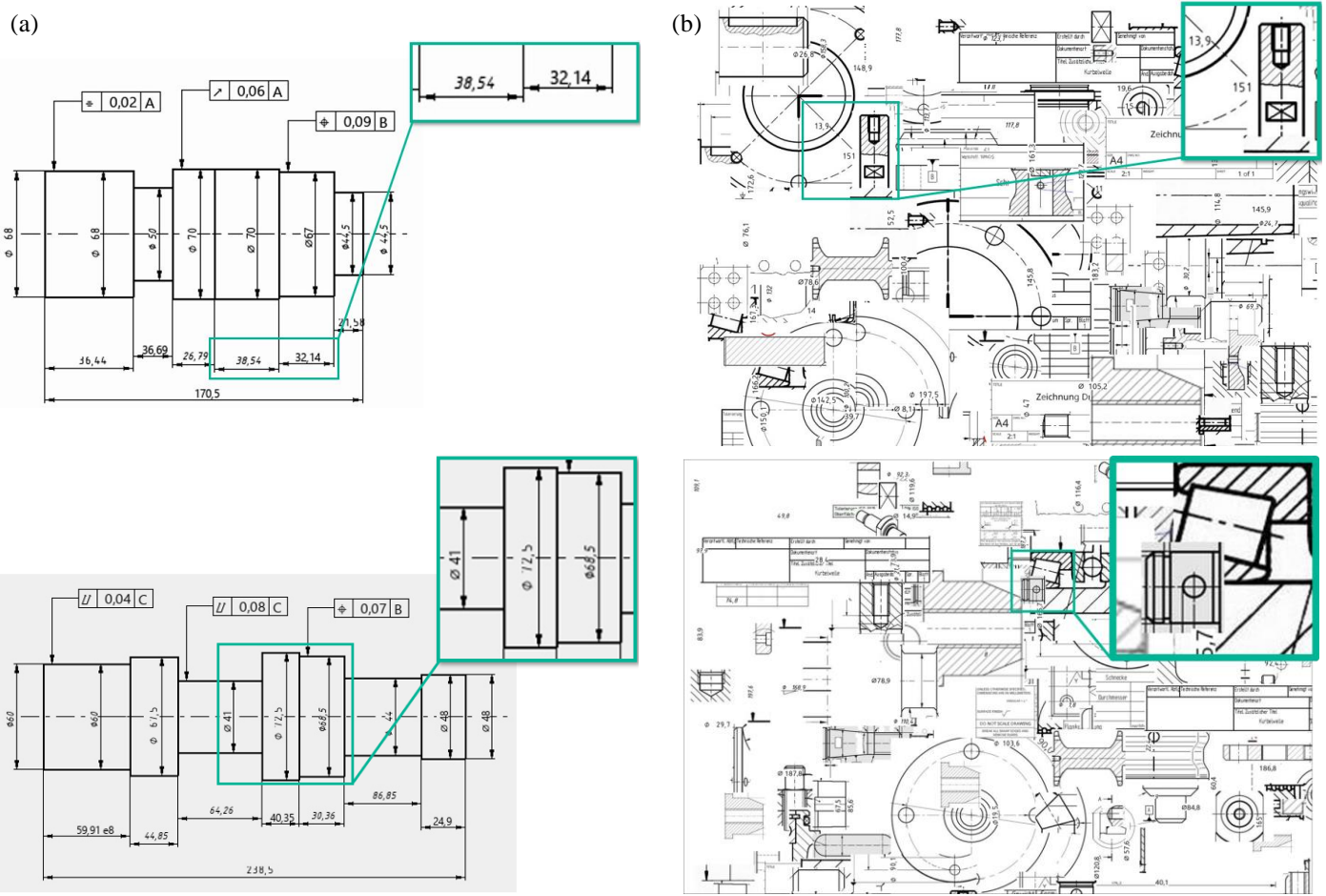

Figure 2: (a) Image type 1: The images show geometries and labels as they may appear on real drawings. The light-gray color of the lower image results from the augmentation of the image. (b) Image type 2: Images contain random elements from 9 real technical drawings to distinguish relevant dimensions and tolerances from background, e.g. bearings and hatchings. The elements are randomly placed, such that intersections of these elements result as well. Best viewed on screen.

### 3.3 Text Recognition

In a next step, the bounding boxes recognized by the Faster-RCNN model are passed to the Text Recognition model to decipher the text inside the bounding boxes. This is the Keras-OCR model that has been used in other use cases [12]. This model has the advantage over other models such as Tesseract OCR [13] that it allows custom training. This makes it possible to recognize special characters such as the diameter symbol and other special characters found on technical drawings. This OCR model normally consists of two parts. A detector and a recognizer. Since there is already a detector in the pipeline, only the recognizer of the OCR model is used, since the required bounding boxes are provided by the object detection model already described. The recognizer was trained using the image type 3 described in Chapter 3.1. The number of images for training the recognizer is 30,000.

### 3.4 Evaluation Metrics

To evaluate the approach, metrics are needed that describe the quality of the object detection model. On the other hand, it must be possible to assess the quality of the recognition. To check the object detection model, the metric Mean Average Precision (mAP), based on [14], is used. This metric is chosen because it not only provides information about how well an object is classified, but also on how well it is localized. Furthermore, this metric is used to evaluate how well the model can be trained with purely artificial images, which in turn provides information about the quality of the images generated by the generator. The mAP is considered at a threshold of Intersection over Union of 0.5. To evaluate the ability to decipher text, the ability of the text recognition part is then evaluated with the Character Error Rate (CER), based on [15]. This is a variation of the word error rate, which indicates the percentage of words in a word sequence that were incorrectly predicted. The CER was chosen because this paper only deals with single-digit sequences that are to be recognized.

## 4. Experiments & Result

To obtain information about the capability of the model, it is analyzed using the metrics described, and these are then compared with an existing out-of-the-box model. The widely used Keras-OCR model, which has already achieved good results in other fields of application, serves as a comparison [12]. Besides the development of an approach to the recognition of text on technical drawings, the aim of the paper is the verification of the effectiveness of images generated artificially by means of a domain knowledge-based generator. To verify the ability of the generator and its artificially generated images, the detection part of the developed approach is trained with different data sets and compared in the next step. The data set generated by the generator consist of a combination of images of image types 1 and 2. The data sets are composed as follows:

- Data set 1: 7000 generated images
- Data set 2: 5000 generated images
- Data set 3: 2000 generated images
- Data set 4: 9 real technical drawings

This comparison provides information as to whether the successful use of a generator for the creation of artificial images is possible compared to limited real drawings. In the comparison of the approach with Keras-OCR, the dataset with 7000 artificial images is used for training. Here, the datasets consist of 70% image type 1 and 30% image type 2. Since the out-of-the-box Keras-OCR is used for comparison, no further training is required.

### 4.1 Results

Figure 3 shows the mAP results at a threshold of 0.5 for the validation data. The specifications described in Section 3.1 were applied. The mAP curves of the model trained with 2000, 5000, 7000 artificial drawings and 9 real engineering drawings are shown. Due to the validation against another 9 real engineering drawings, no train-validation split of the generated dataset is necessary.

Figure 3. mAP comparison between datasets.

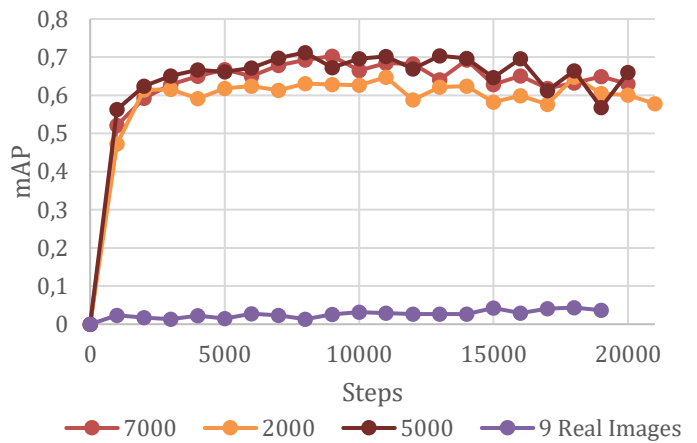

Figure 3 shows that over the whole training process, the model trained with few real technical drawings is far behind. Thus, it is confirmed that the model with few available technical drawings is not functional. The models trained using the generator show much better results. The comparison of the three models reveals that the model trained with 2000 images performs the worst of the 3 models. The curves of the models with 5000 and 7000 images perform similarly. However, as the number of images generated increases, the mAP improves less and less. This confirms the successful use of an artificial image generator in the use case of technical drawings. After testing the detector and the ability of the generator using different data sets, the comparison of the approach with the out-of-the-box Keras-OCR model follows. The result can be seen in Table 1. When comparing the different models, the approach trained using the generator shows the best detection rate (DR). This is closely followed by the Keras-OCR model. The approach trained using 9 real drawings performs significantly worse than the other two models with a DR of 4.67%. For text recognition, the approach with generated data achieves the best result compared to the other models with a margin of 33%. Almost 80% of the digits were recognized correctly. When evaluating the text recognition of the approach with real drawings, the calculated value can vary strongly due to the small number of recognized measures (8). Table 1 shows that the approach trained with only a few technical drawings does not reliably recognize text. This again emphasizes the use of the generator when data availability is limited. In the following Figures 4, 5, and 6, the results of the individual models are illustrated based on an example drawing [16]. Figure 5 shows the approach that was trained using the generator. It can be seen that all objects were detected and correctly classified for both the "horizontal dimensions" and "vertical dimensions" classes. Also, with the class of the "form and position tolerances", all objects are recognized except for one. In Figure 6, with the approach trained using 9 real drawings, no object is detected correctly except for one object of the class "dimensions horizontal". Figure 4 shows the result of the Keras-OCR model. Here, it can be seen that horizontal dimensions can be recognized. However, it is evident that there are clear deficits in the recognition of vertical texts in the proximity of other drawing elements. Also, only the size of the tolerances can be localized correctly, but not the type of tolerance.

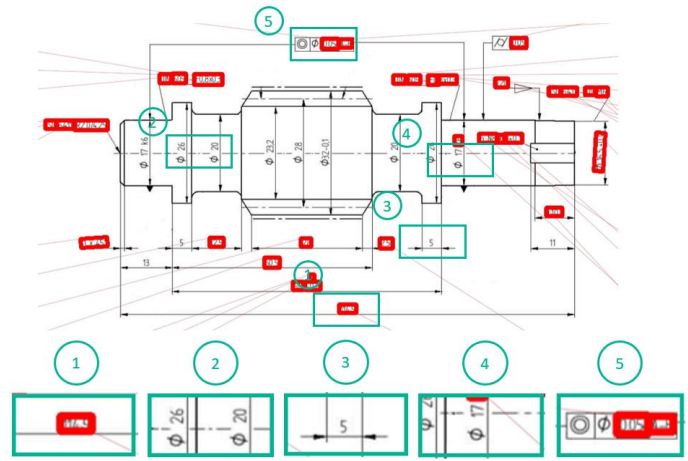

Figure 4. Result of the out-of-the-box Keras-OCR model. Best viewed on screen.

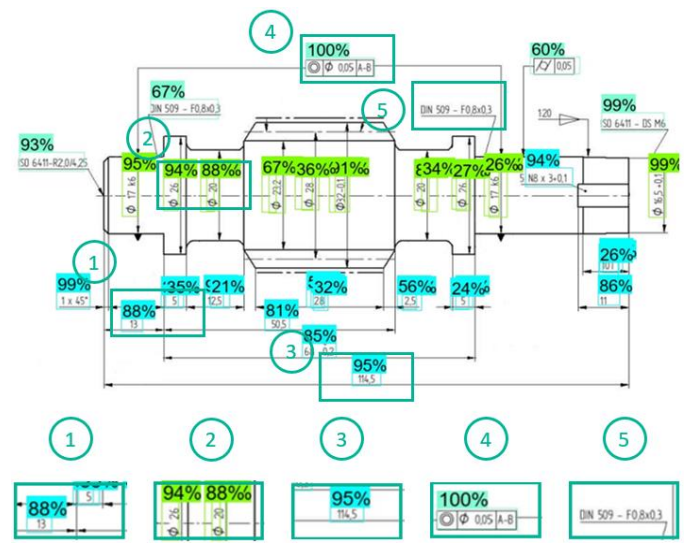

Figure 5. Result of the approach, which was trained with the help of the generator. Best viewed on screen.

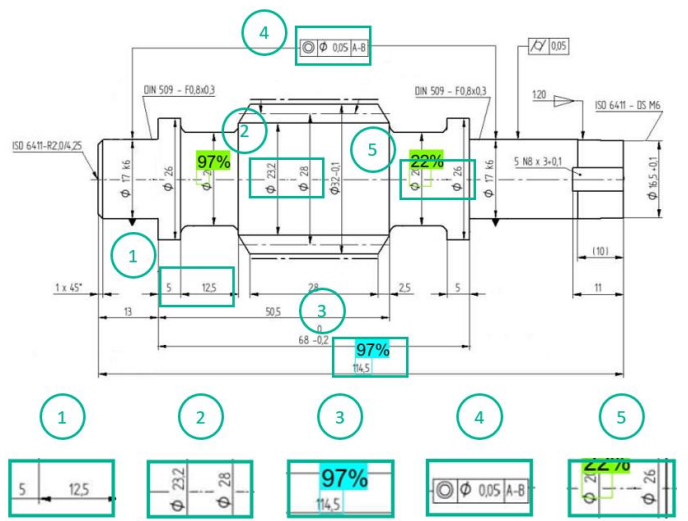

Figure 6. Result of the approach which was trained with 9 real drawings. Best viewed on screen.

Table 1. Comparison approach and Keras-OCR in validation with real technical drawings.

| Model | No. of objects | Detected | Detection rate | Correct text recognition |
|---|---|---|---|---|
| Approach w. generated images | 171 | 140 | 81.87% | 79.33% |
| Approach w. few real drawings | 171 | 8 | 4.67% | 47.91% |
| Out-of-the-box Keras-OCR | 171 | 132 | 77.19% | 46.75% |

Note: Correct Text Recognition results from: 1-CER

## 5. Conclusion

As shown in Figures 4, 5, and 6, the Keras-OCR model has difficulties in recognizing vertical text in the environment of technical drawings as well as in recognizing symbols. These weaknesses of the state-of-the-art model could be solved using the developed approach together with a generator which generated artificial drawings. In addition, the research showed that the application of a generator in the environment of technical drawings is possible especially in the case of a limited amount of available data. The generator enables better text detection on technical drawings compared to existing models. Significantly better results were also achieved in text recognition using the presented approach, especially with the ability to recognize special characters and recognition with varying text orientation. In summary, a functional approach was developed that can be trained based on artificial data generated by a generator. Thus, by integrating domain knowledge, it is possible to reliably recognize text in engineering drawings despite the presence of only small amounts of real-world data. While it is potentially possible to train the Keras-OCR model with the generated data, the limitation is that only characters present in Unicode can be trained. Tolerances in their entirety (type, tolerance size, and reference) cannot be trained with Keras-OCR, which makes it unsuitable for technical drawings.

**References**

[1] H. Fujisawa, "A View on the Past and Future of Character and Document Recognition," Central Research Laboratory, Tokio, 2015.

[2] B. Epshtein, E. Ofek and Y. Wexler, "Detecting Text in Natural Scenes with Stroke Width," IEEE, 2010.

[3] J. Matas, O. Chum, M. Urban and T. Pajdla, "Robust Wide Baseline Stereo from Maximally Stable Extremal Regions," British Machine Vision Conference, 2002.

[4] R. Girshick, "Fast R-CNN," International Conference on Computer Vision, Cambridge, 2015.

[5] Z. Cheng, F. Bai, Y. Xu, G. Zheng, S. Pu and S. Zhou, "Focusing Attention: Towards Accurate Text Recognition in Natural Images," Shanghai, 2017.

[6] C. Zhang, W. Ding, G. Peng, F. Fu and W. Wang, "Street View Text Recognition With Deep Learning for Urban Scene Understanding in Intelligent Transportation Systems," in *IEEE*, 2020.

[7] E. Elyan, L. Jamieson and A. Ali-Gombe, "Deep learning for symbols detection and classification in engineering drawings," in *Neural Networks*, Aberdeen, Elsevier, 2020, pp. 91-102.

[8] L. Jamieson, C. Moreno-Garcia and E. Elyan, "Deep learning for text detection and recognition," Institute of Electrical and Electronics Engineers, 2020.

[9] T. Nguyen, L. Van Pham, C. Nguyen and V. Van Nguyen, "Object Detection and Text Recognition in Large-scale Technical Drawings," in *Proceedings of the 10th International Conference on Pattern Recognition Applications and Methods*, 2021.

[10] S. Ren, K. He, R. Girshick and J. Sun, "Towards Real-Time Object Detection," 2016.

[11] Tensorflow, "GitHub," 2022. [Online]. Available: https://github.com/tensorflow/models. [Accessed 23 March 2022].

[12] Keras, "GitHub," [Online]. Available: https://github.com/faustomorales/keras-ocr. [Accessed 28 March 2022].

[13] R. Smith, A. Abdulkader, R. Antonova and N. Beato, "GitHub," [Online]. Available: https://github.com/tesseract-ocr/tesseract#about. [Accessed 30 March 2022].

[14] M. Everingham, S. M. A. Eslami, L. Van Gool, C. K. I. Williams, J. Winn and A. Zisserman, "The PASCAL Visual Object Classes Challenge: A Retrospective," International Journal of Computer Vision, 2015.

[15] Y. Park, S. Patwardhan, K. Visweswariah and S. Gates, "An Empirical Analysis of Word Error Rate and Keyword Error Rate," in *Proceedings of the International Conference on Spoken Language Processing*, Brisbane, 2008.

[16] T. Hartmann, "Schneckenwelle," 2006.